# Beyond Detection - Orchestrating Human-Robot-Robot Assistance via an Internet of Robotic Things Paradigm

Joseph Hunt[1], Koyo Fujii[1], Aly Magassouba[1], Praminda Caleb-Solly[1][0000-0001-8821-0464]

[1] School of Computer Science, University of Nottingham, UK
`praminda.caleb-solly@nottingham.ac.uk`

**Abstract.** Hospital patient falls remain a critical and costly challenge worldwide. While conventional fall prevention systems typically rely on post-fall detection or reactive alerts, they also often suffer from high false positive rates and fail to address the underlying patient needs that lead to bed-exit attempts. This paper presents a novel system architecture that leverages the Internet of Robotic Things (IoRT) to orchestrate human-robot-robot interaction for proactive and personalized patient assistance. The system integrates a privacy-preserving thermal sensing model capable of real-time bed-exit prediction, with two coordinated robotic agents that respond dynamically based on predicted intent and patient input. This orchestrated response could not only reduce fall risk but also attend to the patient's underlying motivations for movement, such as thirst, discomfort, or the need for assistance, before a hazardous situation arises. Our contributions with this pilot study are threefold: (1) a modular IoRT-based framework enabling distributed sensing, prediction, and multi-robot coordination; (2) a demonstration of low-resolution thermal sensing for accurate, privacy-preserving pre-emptive bed-exit detection; and (3) results from a user study and systematic error analysis that inform the design of situationally aware, multi-agent interactions in hospital settings. The findings highlight how interactive and connected robotic systems can move beyond passive monitoring to deliver timely, meaningful assistance, empowering safer, more responsive care environments.

## 1   Introduction

Incidents such as patient falls are common in hospitals and can lead to further injury, extended stays and financial burden [1]. When surveyed, older patients, reported not wanting to be a burden to healthcare staff and that the act of receiving help causes embarrassment when citing reasons for leaving their bed without assistance [2]. Reliable real-time bed-exit prediction, together with a timely intervention that responds and interacts with the patient to offer support could help to mitigate bed-exit as a reason for falls in hospitals.

Our aim in this research is to develop a cost-effective robotic system with a socially assistive robot (SAR) that responds to events from a thermal sensing bed exit prediction system and determines the patient's needs through interaction. The SAR investigates their underlying motivation for wanting to leave the bed, such as thirst, discomfort, or the need for assistance, before a hazardous situation arises. To demonstrate our system, we piloted a use-case which involved the delivery of items to the patient's bedside, following an interaction where a SARf acquired and communicated their request to another physically assistive robot arm. Our Internet of Robotic Things (IoRT) solution comprises the temi SAR [3], with a separate fixed Kinova Gen3 lite arm [4] with pick and place functionality, mounted on a table.

In this paper we present our findings from a pilot study with participants where this realistic assistive task was conducted in the context of a mock hospital room setting.

The contributions of our study include a modular IoRT-based framework with sensing via low-resolution thermal sensing for accurate, privacy-preserving pre-emptive bed-exit prediction, and multi-robot coordination. The results from this study and the systematic error analysis we conducted can inform the design of better context aware, multi-agent robotic systems. Our paper highlights how interactive and connected robotic systems can move beyond passive monitoring to deliver timely, meaningful assistance, empowering safer, more modular and responsive care environments.

## 2      Related Work

Current methods of detecting a hospital patient leaving a bed include using pressure pads, wearable devices, and infrared sensors. Pressure pads are placed under the user and produce an alert when they no longer detect the patient's presence. These pressure pads are often combined with infrared beams to create 'dual sensing.' A review paper noted how a dual system alarm can have a sensitivity of nearly 100%, with a positive predictive value of 68% but can generate from 16% to 31% "nuisance alarms" [5].

Robots, and more specifically, the temi robot this study used, are showing potential for interacting with patients on hospital wards. A 2023 study into the use of temi robots on isolation wards demonstrated their ability to provide communication between patients and nursing staff using video calls and voice commands. Participants gave this robot a mean difficulty rating of 1.28/5 (lower is better) [6].

The IoRT is a proposed extension of the Internet of Things to include communication between robotics systems and IoT systems [7], [8]. This integration allows for communication between mobile robots, sensors and machine learning algorithms to bring enhanced intelligence to IoT compatible devices. The concept also encompasses communication technologies across multiple layers such as the Message Queuing Telemetry Transport (MQTT) [9] protocol on the application layer to facilitate machine-to-machine communication.

This lightweight, publish-subscribe based messaging protocol which is commonly used for IoT applications is designed for resource-constrained devices and low-bandwidth, high-latency, or unreliable networks. Other studies have also been working towards cooperation of heterogeneous robotic systems [10] also analysing error between robots and users given a sequential task [11], [12].

## 3    Approach

We have designed an IoRT system composed of a social robot (temi), and Kinova Gen3 Lite robotic arm, connected to a bed-exit prediction system. Our setup is proposed as a modular integrated solution to grasping and transporting objects by using multiple independent robotic systems that communicate with each other and the bed-exit prediction system through MQTT. Our selection of platforms aims to show how we might produce a workable solution by combining robots, whereby dividing complex tasks among specialised robots, results in more efficiency. Furthermore, the modularity of the system can improve real-world usability, allowing it to be tailored to specific scenarios. Environments such as clinical settings often have confined areas, which are not conducive to large mobile robots with arms. Additionally in hospitals avoiding cross-contamination is important [13] so a system with several local and smaller robots could be more pragmatic. Our approach using several smaller heterogeneous robot systems cooperating in an IoRT framework could also reduce the total cost of the system. We have designed the proposed system with the potential to be highly scalable, where individual robots could be swapped out and more systems integrated without incurring significant development costs and downtime. The system architecture is shown in Fig 1 and each system sub-component is discussed in the following sections.

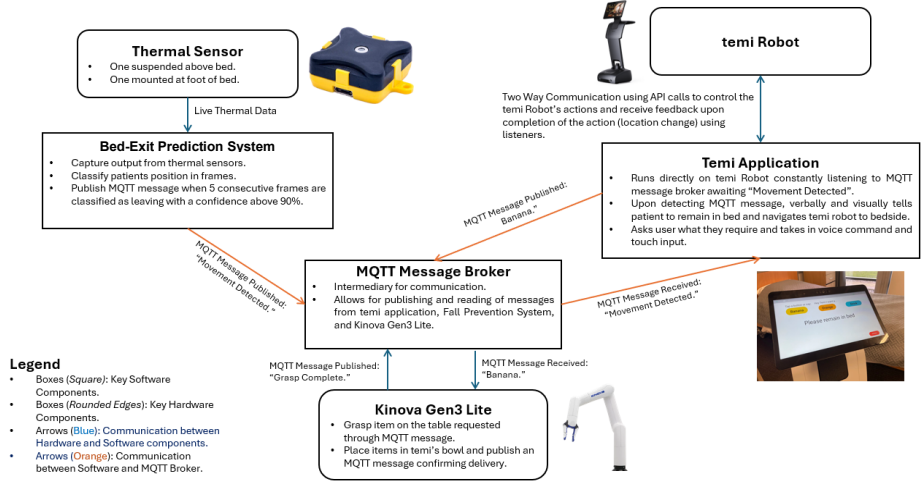

Fig 1. Components within our IoRT system and system architecture.

### 3.1    Bed-Exit Prediction System

Two Terabee TeraRanger Evo thermal cameras [14] were positioned in a fixed locations as shown in Fig 2c, at the foot of a hospital bed and above it. The locations were selected to avoid being obtrusive and not interfere with changing of bed linen or the patient. These sensors provide a 32x32 pixel resolution (Fig. 2) image at a frame rate of 14 Hz allowing for accurate classification whilst conserving the patient privacy.

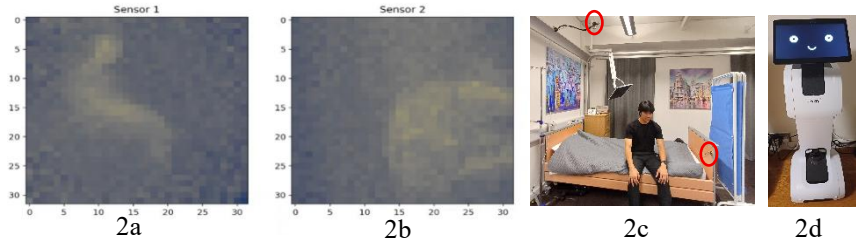

2a    2b    2c    2d

Fig 2. Images from thermal imaging sensors from the side and top cameras (circled in red) showing a person seated with their legs at the side of the bed, and temi robot.

### 3.2    temi – Social Interaction and Transport Robot

The temi (Fig 2d) is a SAR with mapping, navigation and collision avoidance features which are crucial for deployment in busy environments such as a hospital ward. It also has a 13.3" capacitive touch screen for a graphical user interface or animated face, as well as built-in natural language processing and speech synthesis, with a microphone and speakers, making it an ideal platform for social interaction.

For this study, a Kotlin android application was developed and deployed on the temi robot. Our application enabled the robot to navigate autonomously between key locations based on events triggered from the bed-exit prediction system. We also implemented multi-modal interaction features; the robot uses speech to instruct the "patient" to remain in their hospital bed when an event from the bed-exit system is detected and displays the same text on the integrated touch screen. When the robot reaches the patient's bedside, the temi screen displays an array of buttons labelled with different items that the patient can request to be fetched to them. The patient can also use speech to verbally articulate their request. Based on the location of the requested item, the temi robot navigates to the appropriate pickup position where the bowl it is carrying on a tray can be loaded with the selected item by the Kinova Gen3 Lite which completes a pick and place task. The pick and place task, and the action of the temi returning to the bedside with the item, is coordinated and synchronised via MQTT messages.

### 3.3    Kinova Gen3 Lite Manipulation robot

A robot arm, a Kinova Gen3 Lite with a RealSense camera, was used to perform the pick and place manipulation task using MoveIt, a robotic manipulation platform for ROS [15]. The arm was mounted at the edge of a table with a range of items on it. The system can be scaled by using lower cost arms mounted in different locations, and the item could also potentially be requested from hospital staff who might be elsewhere. We utilised a pre-trained state-of-the-art image segmentation model, Early Vision-Language Fusion for Text-Prompted Segment Anything Model [16] to detect the requested object based on a language prompt. The name of item to be picked up is passed to the Kinova arm sub-system via MQTT from the temi as per the user's request to the temi.

### 3.4 Communication protocol

We used the Message Queuing Telemetry Transport (MQTT) protocol to establish the communication between the different elements of the IoRT system. One advantage of MQTT compared to other protocols such as HTTP, is the low bandwidth [17] making it suitable for resource-constrained hospital environments. MQTT is also low-latency [18] which is important for minimising the time it takes for temi to instruct the patient not to leave and bed and reach the patient's bedside after a bed-exit is event triggered.

## 4 Experiment and Results

### 4.1 Experiment Design

Sixteen participants (11M & 5F, mean age 22) participated in our study with the IoRT system. All participants were students at the University of Nottingham (UoN) who provided ethics approval for the study. The study took place in a mock hospital ward room which housed a bed, the temi, and Kinova Gen3 robot, and a table. On the table, there were several items, such as a water bottle and different fruit. The participants' interaction with the system was video recorded and they were asked to fill a user experience questionnaire at the end of the trial.

The experiment was divided into two parts. The first part was to test to sensitivity and accuracy of the bed-exit prediction model. Each participant was instructed to lie down on the bed and attempt to leave the bed whenever they wanted. When the bed-exit prediction system detected an attempt to leave the bed, an alarm would sound. At this point, the participant was instructed to get back onto the bed and restart the scenario. To test the robustness of the prediction model, this trial was repeated 15 times; however, the participant was told to select 5 of these trials at random and attempt to "trick" the system into thinking they were getting out of the bed when they did not intend to.

In the second part of the experiment, which was repeated three times, the user experience and interaction with the temi robot was tested. This time, when a participant was detected leaving the bed, instead of the sound alarm, the temi robot would instruct the participant to remain in bed and start moving to the person's bedside and ask the participant if there was anything it could do for them. The participant could then select an item to fetch, either through voice or through the touch screen. The robot then navigated to the pickup location. At this point, the Kinova Gen3 Lite picked up the requested item from the table and placed it into temi's bowl. The temi then returned to the participant's bedside before turning around and instructing them to take the item.

During all stages of the robotic interaction, the temi audibly informed the participant of its current actions and intentions, including audibly "telling" the Kinova arm what item had been requested as it published the MQTT message. This was done to ensure transparency and keep the participants aware of what was happening. Participants were purposely given no instruction or training on how to interact with the temi robot and were told there was no right or wrong way to interact, as we were also interested in understanding what their pre-conceived assumptions about the system operation would be, and also to understand the likely errors that might ensue.

### 4.2  User Experience Questionnaire Results

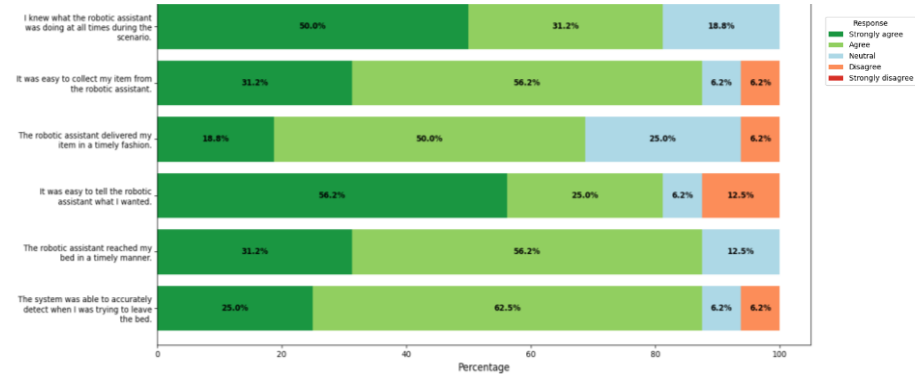

Fig 3. Responses to User Experience Questionnaire

The questionnaire contained Likert-scale user experience questions designed to evaluate participants' perceptions of the system as shown in Fig 3. It also included two open-ended questions that asked what participants felt the system did well and what it did not do well. When asked what they liked, participants praised the robot's human interaction capabilities and its non-threatening, friendly presence. Comments included: "The voice of the robot is soft and has a gentle tone which could be seen as non-confrontational," and "temi is a fairly non-scary robot, which is definitely important." Another participant noted, "The voice of the robot was not intimidating and comforting and helped to dispel any nervousness about it." Participants also complimented the clarity of the user interface and the ease of interaction through multi-modal input. One participant stated, "I think the user interface on the robot was easy to use and navigate—there was no possible confusion about how to interact with it when it came to my bedside." Another remarked, "The user interface is clear, and system is clean and easy to operate." On the other hand, when asked what they did not like, many participants felt that the speed of the system could be improved. Comments included: "The fetching of the items could be quicker," and "It took a long time to get the item once the robot had gone to get it (the robotic arm took a long time)." Additionally, several participants pointed out issues with the robot's navigation. One participant observed, "The robot did not face me when I tried to leave the bed at the foot of it. This gave me the impression that the robot did not actually know where I was and instead positioned itself in a general position. The interaction became less personal at that point."

### 4.3  Interaction Results

The bed-exit prediction system had a classification accuracy of 79.65% on unseen data, with recall of 1, the misclassifications were all false alarms. This could be annoying if it meant that a nurse had to be called out, but with our system, this could reduce pressure on staff. Participants preferred using touch input to tell the temi robot what item they wanted. Touch input was used by participants 29 times whereas voice input was only used 15 times. Participants began to use the touch screen less in trials two and three and instead, increasing their use of voice to tell the temi robot what they wanted.

### 4.4 Analysis of interaction errors

To analyse the errors made by both the participants and the robots involved in the experiment, an error taxonomy was created based on previous research [19], [20]. A total of 10 HRI error types were considered. To support consistent and efficient error logging, a simple GUI was developed with all the error categories, and the interaction modality. A sample of the videos were randomly selected and reviewed by other members of the study team to check for inter-rater variability. The error types included Social errors (misunderstanding the user, insufficient communicative function); Safety-related errors (human errors: procedure, intrusion, operation and situation awareness) and Robot errors (system failures, safeguarding failures, operational errors, design flaws). Operational errors were the most common for both interaction modalities with 66 instances, mostly related to the temi attempting to avoid the participants' legs and nearby armchair and failing to get close enough to the participant. Common human errors were intrusion errors (4 instances) where participants intentionally or unintentionally interrupted the flow of operations by using the trigger word or saying something else. Insufficient communicative function was the most common social error (4 instances), due to errors in synchronising listening and responding to participants. The rarest errors were system failures and design flaws, with only 1 instance each. Categorising errors using this taxonomy helps to prioritise improvements that need to be made to the platform before further use.

## 5 Conclusions

This paper proposes an Internet of Robotics Things paradigm that utilises two different types of robots coordinating their tasks for pro-active assistance in hospital environments. Our system was evaluated in a realistic user study which has helped to highlight system weaknesses to address before a real-world deployment. Based on this study, we have noted several areas for further study and improvements. These include:
- Studying how human adaptation can be exploited to decrease HRI errors over time.
- Personalising the system behaviour and interaction modalities based on the condition of the patient who is being supported. We will consider constructing several user models which can be used to determine the best approach.

Our next phase will involve a larger study with real patients in a clinical setting.

## References

[1] R. T. Morello *et al.*, 'The extra resource burden of in-hospital falls: a cost of falls study', *Medical Journal of Australia*, vol. 203, no. 9, pp. 367–367, 2015, 10.5694/mja15.00296.
[2] T. P. Haines, D. A. Lee, B. O'Connell, F. McDermott, and T. Hoffmann, 'Why do hospitalized older adults take risks that may lead to falls?', *Health Expectations*, vol. 18, no. 2, pp. 233–249, Apr. 2015, doi: 10.1111/hex.12026.

[3]   'Get temi the personal robot for your business | robotemi.com', temi robot. Accessed: Apr. 01, 2025. [Online]. Available: https://www.robotemi.com/product/temi/

[4]   'Discover our Gen3 lite robot', Kinova. Accessed: Apr. 01, 2025. [Online]. Available: https://www.kinovarobotics.com/product/gen3-lite-robots

[5]   M. Oh-Park, T. Doan, C. Dohle, V. Vermiglio-Kohn, and A. Abdou, 'Technology Utilization in Fall Prevention', *Physical Medicine & Rehab*, vol. 100, no. 1, p. 92, Jan. 2021

[6]   H. J. Yoo, E. H. Kim, and H. Lee, 'Mobile robots for isolation-room hospital settings: A scenario-based preliminary study', *Computational and Structural Biotechnology Journal*, vol. 24, pp. 237–246, Dec. 2024, doi: 10.1016/j.csbj.2024.03.001.

[7]   'Internet of Robotic Things: Driving Intelligent Robotics of Future - Concept, Architecture, Applications and Technologies | IEEE Conference Publication | IEEE Xplore'. Accessed: Mar. 18, 2025. [Online]. Available: https://ieeexplore.ieee.org/abstract/document/8611051

[8]   M. Sandhu, D. Silvera-Tawil, P. Borges, Q. Zhang, and B. Kusy, 'Internet of robotic things for independent living: Critical analysis and future directions', *Internet of Things*, vol. 25, p. 101120, Apr. 2024, doi: 10.1016/j.iot.2024.101120.

[9]   D. Soni and A. Makwana, *A Survey On Mqtt: A Protocol Of Internet Of Things(Iot)*. 2017.

[10]  J. Kiener and O. von Stryk, 'Towards cooperation of heterogeneous, autonomous robots: A case study of humanoid and wheeled robots', *Robotics and Autonomous Systems*, vol. 58, no. 7, pp. 921–929, Jul. 2010, doi: 10.1016/j.robot.2010.03.013.

[11]  S. Shin, Y. Kwon, Y. Lim, and S. S. Kwak, 'User Perception of the Robot's Error in Heterogeneous Multi-robot System Performing Sequential Cooperative Task', in *Social Robotics*, A. A. Ali, J.-J. Cabibihan, N. Meskin, S. Rossi, W. Jiang, H. He, and S. S. Ge, Eds., Singapore: Springer Nature, 2024, pp. 322–332. doi: 10.1007/978-981-99-8718-4_28.

[12]  S. Shin and S. S. Kwak, 'Do Hierarchies in a Robot Team Impact the Service Evaluation by Users?', *2023 IEEE/RSJ Int Conf on Intelligent Robots and Systems*, pp. 3983–3990.

[13]  C. Lim, M.-Y. Lee, and S.-W. Kim, 'Recent development of medical service robots in Republic of Korea and their field demonstration cases in the clinical setting', *JMST Adv.*, vol. 6, no. 4, pp. 395–401, Dec. 2024, doi: 10.1007/s42791-024-00092-y.

[14]  'TeraRanger Evo Thermal Cameras'. Accessed: Mar. 25, 2025. [Online]. Available: https://www.mouser.co.uk/terabee-thermal-cameras

[15]  D. T. Coleman, I. A. Sucan, S. Chitta, and N. Correll, 'Reducing the Barrier to Entry of Complex Robotic Software: a MoveIt! Case Study', Italy, 2014.

[16]  Y. Zhang *et al.*, 'EVF-SAM: Early Vision-Language Fusion for Text-Prompted Segment Anything Model', Mar. 10, 2025, *arXiv*: arXiv:2406.20076.

[17]  H. H. Alshammari, 'The internet of things healthcare monitoring system based on MQTT protocol', *Alexandria Eng Journal*, vol. 69, pp. 275–287, Apr. 2023

[18]  A. Gavrilov, M. Bergaliyev, S. Tinyakov, K. Krinkin, and P. Popov, 'Using IoT Protocols in Real-Time Systems: Protocol Analysis and Evaluation of Data Transmission Characteristics', *Com Networks and Comms*, vol. 2022, no. 1, p. 7368691, 2022

[19]  N. Batti *et al.*, 'Improving Human Understanding of Errors Through Enhanced Robot-to-Human Error Reporting'. RSS 2024, Workshop on Robot Execution Failures

[20]  B. H. W. Guo, Y. Zuo, Y. M. Goh, and J.-Y. Lim, 'Errors in Human-Robot Interaction Accidents: A Taxonomy and Network Analysis', *Int conf on construction engineering and project management*, pp. 1088–1095, 2024, doi: 10.6106/ICCEPM.2024.1088.